\documentclass[10pt, conference, compsocconf]{IEEEtran}
\usepackage[pdftex]{graphicx}

\usepackage{amssymb}
\usepackage{booktabs}

\ifCLASSINFOpdf
\else
\fi
\usepackage[cmex10]{amsmath}
\begin{document}
%
\title{Object Class Aware Video Anomaly Detection through Image Translation}


\author{\IEEEauthorblockN{Mohammad Baradaran}
\IEEEauthorblockA{Université Laval\\
Computer Vision and Systems Laboratory (CVSL)\\
Québec City, Québec, Canada\\
Email: mohammad.baradaran.1@ulaval.ca}
\and
\IEEEauthorblockN{Robert Bergevin}
\IEEEauthorblockA{Université Laval\\
Computer Vision and Systems Laboratory (CVSL)\\
Québec City, Québec, Canada\\
Email: robert.bergevin@gel.ulaval.ca}
}


%


\maketitle

\begin{abstract}
Semi-supervised video anomaly detection (VAD) methods formulate the task of anomaly detection as detection of deviations from the learned normal patterns. Previous works in the field (reconstruction or prediction-based methods) suffer from two drawbacks: 1) They focus on low-level features, and they (especially holistic approaches) do not effectively consider the object classes. 2) Object-centric approaches neglect some of the context information (such as location). To tackle these challenges, this paper proposes a novel two-stream object-aware VAD method that learns the normal appearance and motion patterns through image translation tasks. The appearance branch translates the input image to the target semantic segmentation map produced by Mask-RCNN, and the motion branch associates each frame with its expected optical flow magnitude. Any deviation from the expected appearance or motion in the inference stage shows the degree of potential abnormality. We evaluated our proposed method on the ShanghaiTech, UCSD-Ped1, and UCSD-Ped2 datasets and the results show competitive performance compared with state-of-the-art works. Most importantly, the results show that, as significant improvements to previous methods, detections by our method are completely explainable and anomalies are localized accurately in the frames.

\end{abstract}

\begin{IEEEkeywords}
video anomaly detection; deep learning; semantic segmentation; semi-supervised learning

\end{IEEEkeywords}

%
\IEEEpeerreviewmaketitle

\section{Introduction}
\label{sec:intro}

Constant monitoring of the outputs of surveillance cameras by human operators is not practically effective and reasonable. Hence, there is a demand for an intelligent system to automatically analyze the content of videos, localize the suspicious cases, and detect the events of interest for further analysis. Among all massive volumes of video data, abnormal video events are of greater importance for potential security measures. Video anomaly detection (VAD) is about identifying events that do not conform to regular events, or in other words, deviate significantly from the expectations \cite{chandola1}. Anomalies happen rarely, and they can appear in unlimited types \cite{liu1}. Hence, enough labelled abnormal data is not practically available for supervised training. On the other hand, normal data can be collected quickly and in a considerable amount, which can be used for semi-supervised training. Moreover, addressing the problem as a semi-supervised learning task is more compatible with the definition of anomaly detection since, generally, the definition of the anomaly is tied with the concept of normality.

\begin{figure}[ht]
  \centering
   \includegraphics[width=1\linewidth]{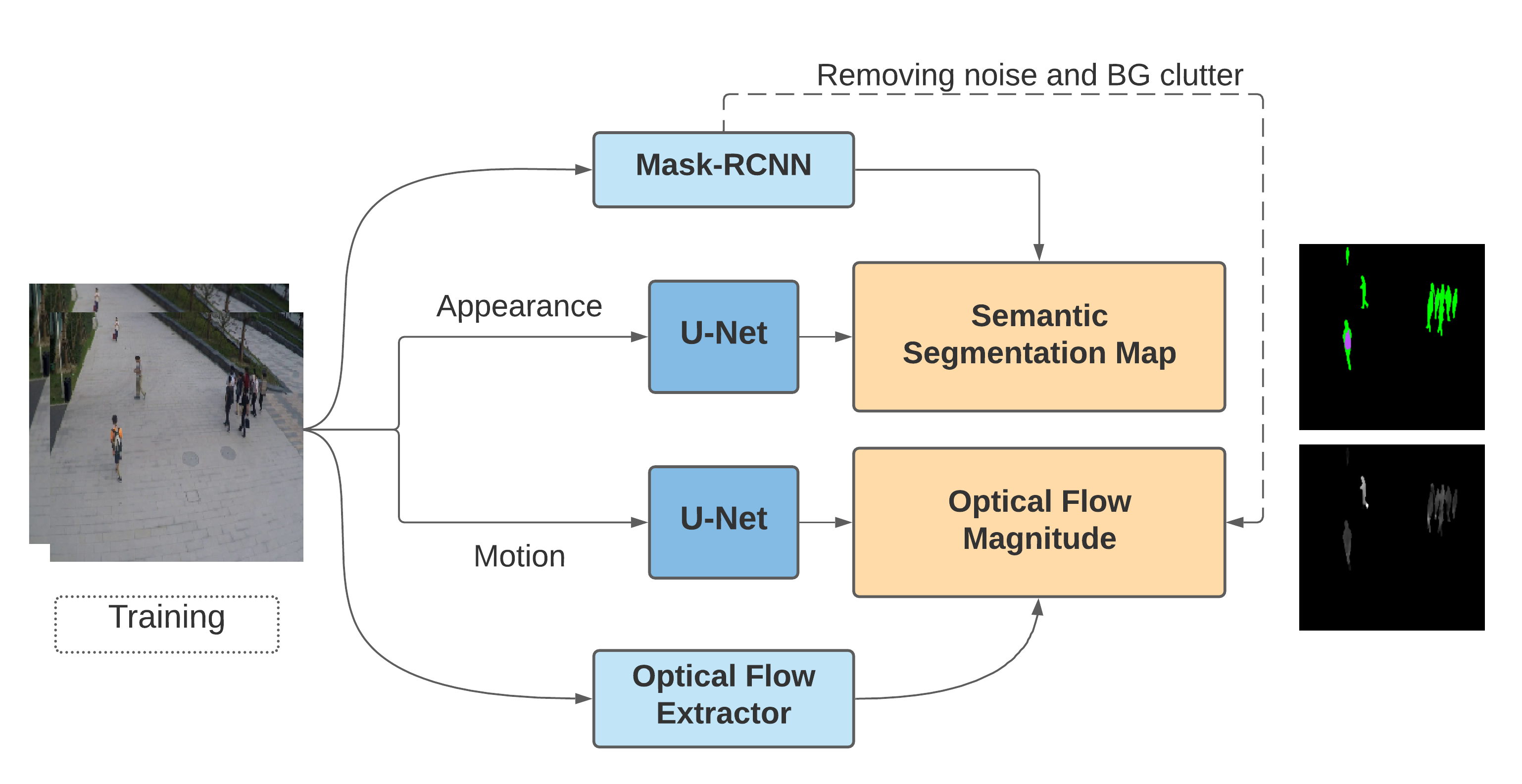}
   \includegraphics[width=1\linewidth]{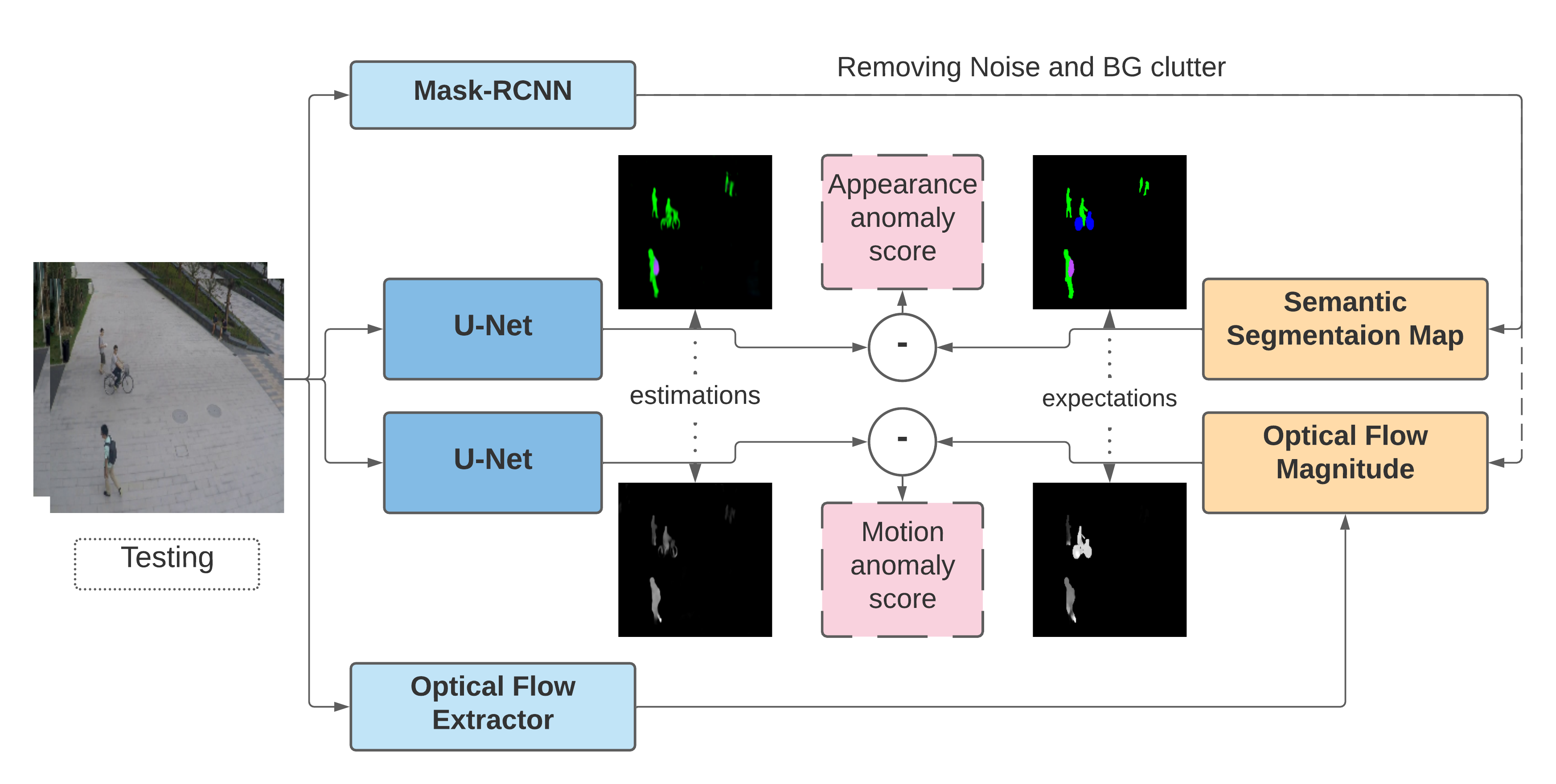}
   \caption{The pipeline of our method. Best viewed in color.}
   \label{fig:model}
\end{figure}

Deep learning (DL) based video anomaly detection methods have achieved significant improvements with respect to their classic counterparts. Previous DL-based holistic approaches \cite{hasan1, chong1, abati1, liu1} generally, concentrate on low-level features to represent normal frames and consequently to identify anomalies. Hence the class of the objects is not considered for the final decision, while the objects' class practically plays the most critical role in defining the anomalies in surveillance cameras from an appearance viewpoint. On the other hand, object-centric approaches \cite{ionesco1}, \cite{doshi1} devote their concentration just to detected objects, to consider the objects better. However, they still learn normality by focusing on low-level features (by reconstructing or predicting the cropped images of the detected objects). Moreover, they neglect important context information (such as the location of the object in the scene) as they crop the object out of the frame. 
A major drawback of the previous methods is the poor explainability of their detections \cite{kasun1}. Even if a frame is accurately identified as an anomaly, it is imperative to make sure that the decision has been made based on an anomaly-related reason (e.g., abnormal object or motion), not an unrelated factor (such as the noise, number of objects, illumination changes, etc.). Importantly, the explainability of a method facilitates the localization of the anomaly. Previous methods have not discussed their method from this point of view. Moreover, re-implementations of different methods show that either they have a considerable number of unexplainable detections or do not have precise attention on anomalous regions.

To address the mentioned issues, we propose a two-stream approach to detect anomalies in video. One branch learns the normal appearances while the second one considers the motion. Unlike previous methods, our proposed method considers the class of the object for anomaly detection instead of low-level features (such as intensity, colour, etc.). Low-level features can not effectively represent anomalies in surveillance videos, and they are more vulnerable to noise and illumination changes. In this branch, we train a U-net (with a resnet34, pre-trained on ImageNet, as encoder) to translate the input frame to its target semantic segmentation map. This formulation helps the network learn normal objects in the frame, considering their classes. In parallel, in the motion branch, an identical network translates the input frame to its optical flow magnitude map. In this way, the network learns to make an association between objects and their regular motions. In summary, our contributions in this paper are as follows:

\begin{itemize}
\itemsep0em
\item We propose a novel two-stream video anomaly detection method that concentrates on the class of the objects instead of focusing on low-level features. This goal is achieved by reformulating the problem as an image translation from the input frame to its pre-calculated semantic segmentation map. To the best of our knowledge, this is the first study to consider the class of the objects for VAD.
\item We formulate the task of learning normal motion as an image translation from the input frame to its target optical flow magnitude map.
\item Our method is explainable, and it identifies a frame as an anomaly based on only anomaly-related activations. Moreover, anomalies are accurately localized in the frames considering the anomaly map activations.
\item Our proposed method outperforms state-of-the-art methods on the Shanghaitech dataset and shows a competitive results on the UCSD datasets.
\end{itemize}

The remainder of this paper is organized as follows: In Section {II}, we review the related literature in video anomaly detection. Section {III} describes our proposed method in detail. Experiments and results are described in Section {IV}, and some conclusions are drawn in Section {V}. 

\section{Related works}
\label{sec:hist}

Video anomaly detection is generally addressed in a semi-supervised manner. Researchers have addressed the problem mainly by formulating the task as a reconstruction task \cite{hasan1, chong1, abati1, samet1, ravanbakhsh1, park1} or a prediction task \cite{liu1, chen1, pankaj1, yu1}. In reconstruction-based approaches, an unsupervised network (usually an Autoencoder, U-net, etc.) is trained to reconstruct the normal frame(s), assuming that abnormal frames would result in a higher reconstruction error. Prediction-based approaches on the other hand get several consecutive frames of a normal clip and are trained to predict the future frame(s), expecting a higher prediction error for an abnormal frame. These methods inherently consider the evolution of the frames to predict the future frame and hence capture motion patterns. To benefit the advantages of both reconstruction and prediction frameworks, \cite{zhao1, morais1} propose a hybrid approach, having both approaches in different branches. 

Reconstruction-based and prediction-based approaches have drawbacks: 1) They generally consider low-level features for prediction and are not aware of the class of the objects effectively \cite{mohammad1,yu1}. 2) The methods that simultaneously model appearance and motion in a single stream fail in considering motion as effectively as appearance.

Unlike mentioned holistic approaches, object-centric approaches \cite{ionesco1} only focus on detected objects. These methods crop the objects out of the frame and learn their appearance and motion individually. Object-centric approaches apply their focus on objects, and hence they can be generalized to different scenes. Moreover, they do not have to deal with computational complexity due to background \cite{pankaj1}. However, they still have two main drawbacks: 1) They crop objects out of the frame, process them individually, and do not consider the location information. 2) Reconstructing or predicting the cropped objects’ images helps the network focus on the objects but does not necessarily guarantee to learn the objects' classes.

Recently, Krzysztof et al. \cite{krzystof1} proposed a method to detect the novelties in images by reconstructing the frame from its semantic segmentation map. Through this framework, the network considers the class of the object. Inspired by this method and \cite{biase1}, we benefit from semantic segmentation in a different way and we formulate the problem as image translation from the original frame to its semantic segmentation map. Moreover, unlike the previous work, which only considers appearance, we add a second branch to capture motion patterns.

Although different strategies have been applied to formulate and consider motion for the VAD, the correspondence between the motion and the object has not been considered effectively. For this purpose, Nguyen et al. \cite{trong1} formulate the problem of learning normal motion as an image translation task. They train a network to translate from a raw frame to its corresponding optical flow map. However, it is confusing for the network to identify the direction of motion for the objects from a single frame. We implemented a similar approach; however, we trained our network to translate the raw frame to the corresponding optical flow magnitude map. Hence, the network only considers the magnitude of the motion and does not get confused by the direction.
\section{Proposed Method}
\label{sec:method}

We propose a deep learning-based semi-supervised video anomaly detection method that takes advantage of two parallel branches to model normal appearance and motion individually and, consequently, to detect anomalies, leveraging image translation. The pipeline of the proposed method is shown in Fig. 1, which illustrates the training and the inference (testing) phases in detail. The details of the proposed method are described in the following sub-sections.

\subsection{The two-stream method}
Our method models normal appearance and motion separately in two different but similar branches. Experiments in \cite{mohammad1} suggest that modelling motion and appearance separately would maintain the effect of each factor on the final decision, unlike the single branch methods, in which one feature may be dominated by the other.

\subsection{The appearance branch}

In the appearance branch, we train a U-net with a pre-trained resnet34 (trained on the ImageNet) as an encoder. The network is trained to learn the translation from the input frame to its target semantic segmentation map. The target semantic segmentation maps of the frames are acquired using a state-of-the-art object segmentation method: the Mask-RCNN. It is assumed that the network learns to identify the objects inside the frame and recognize their object classes through this formulation. In this way, the networks learn to focus on the objects without needing to crop them out of the frame (as is done in object-centric approaches) and identify their classes. 
\subsection{The motion branch}

In the motion branch, the same formulation, as we had for the appearance, has been applied, and an identical U-net is trained to learn translation from the input image to its corresponding optical flow magnitude map. The optical flow magnitude map is a grayscale image that keeps only the magnitude information of the optical flow and removes the direction information (i.e., colours in the optical flow map). Through this formulation, the network learns to associate the objects and their normal motions at different locations. It is worth mentioning that the translation is from a single image to another single image. In other words, we do not use two consecutive frames as an input for the U-net to learn the optical flow calculation. However, consecutive frames are used to create target images, and the network learns the correspondence between each object in the input frame and its pre-calculated motion magnitude. Our experiments showed that the network gets confused about the direction when it tries to learn the translation to the raw optical flow map and does not produce useful results. However, it properly learns the normal motion magnitudes for each object. Fig. 2 illustrates this challenge and its solution.

We developed our holistic approach such that location information is not neglected. This plays a crucial role in scene anomaly detection since, from an appearance viewpoint, generally, each object type is expected to appear in particular regions in the frame (for example, cars are expected to appear on the street, not on the sidewalk). The same rule is applicable to motion as well. Most particularly, our approach is expected to learn different normal motions (even for the same object type) at different locations. For example, pedestrians at a distance produce a lower motion magnitude (regarding their optical flow results) than near the camera. Our experiments show that our proposed networks learns normal motions for each object, considering their locations. 

\begin{figure}[t]
  \centering
   \includegraphics[width=0.8\linewidth]{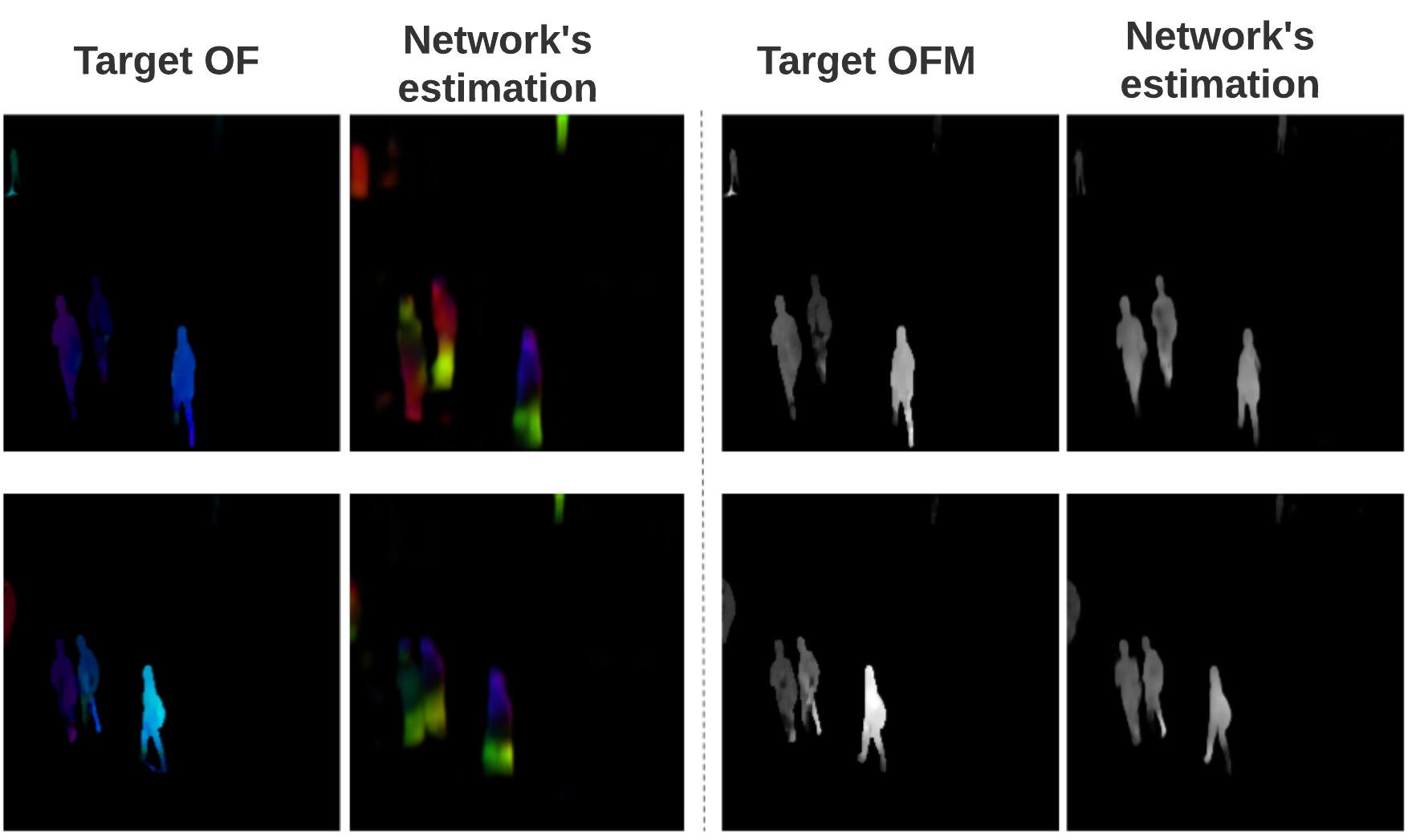}

   \caption{Comparing the results of motion learning for optical flow (Left) and optical flow magnitude (Right). OF and OFM stand for the optical flow and optical flow magnitude, respectively. Best viewed in color.}
   
   \label{fig:trans}
\end{figure}

\subsection{Masking}
To focus on the motion of the objects, we apply the calculated semantic segmentation mask (produced by the Mask-RCNN) on the calculated target optical flow map. In this way, the background motion is suppressed which helps the network focus and learn the motion of the detected objects. Fig. 3 shows samples of noisy raw optical flow map, the corresponding segmentation mask and the final masked optical flow map, which contains only the motion of the objects.

\begin{figure}[t]
  \centering
   \includegraphics[width=0.55\linewidth]{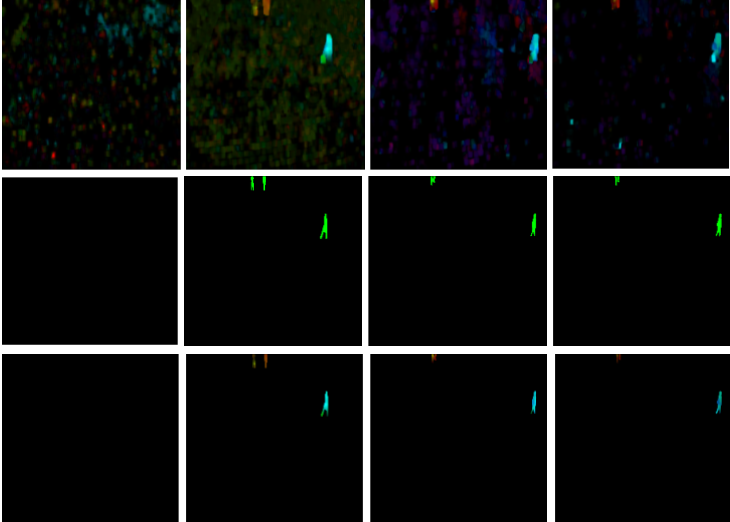}

   \caption{Optical flow masking step. Up: Extracted optical flow. Middle: Segmentation map. Bottom: Masked optical flow map.}
   \label{fig:post}
\end{figure}

\subsection{Training}
We train networks of the appearance and motion branches to minimize the difference between their target images (T) and their outputs $(F_{W} (I_{t}))$. Hence, the optimization follows the following target function (1):
\begin{equation}
Argmin_{w} ( diff (T(I_{t}) , F_{w}(I_{t}) )  
\label{eq:e1}
\end{equation}
Where diff denotes the function that calculates the difference between $T$ and $F_{w}$. $T$ and $F_{w}$ represent the target image for the input I (at time t) and function of the network (U-net), respectively. It is worth mentioning that, inspired by \cite{trong1}, we leverage the L1 loss for the optical flow since it does not magnify the effect of the noise. For the appearance network we leveraged the L2 loss, to calculate the difference between the output and the target image. Moreover, Yang et al. \cite{yang1} believe that using patch-level losses for training forces the network to focus on different regions equally and not prioritize background. Hence, we benefit from the patch-level loss which computes the MSE or MAE losses for imagined patches in the frame (Fig. 4) and select the maximum patch loss as the final loss of the frame. The patch-based loss is calculated as below (2), (3) for the appearance and motion branches:
\begin{equation}
L_{ss}=Max(L_{ss1},L_{ss2},..L_{ssk})
\end{equation}
\begin{equation}
L_{of}=Max(L_{of1},L_{of2}, .. L_{ofk})
\end{equation}

Where:
\begin{equation}
L_{ssi} = {\| Pi (f1_{w}(I_{t})) - Pi (SM( I_{t}))\|}^2
\end{equation}
\begin{equation}
L_{ofi} = \| Pi (f2_{w}(I_{t})) - Pi (OFM(I_{t-1}, I_{t}))\|
\end{equation}

Pi selects the ${i_{th}}$ patch in the frame (as shown in Fig. 4). $L_{ssi}$ and $L_{ofi}$ show the loss of $i_{th}$ patch (out of the k produced patches) for appearance and the motion, while $L_{ss}$ and $L_{of}$ stand for the final loss for the appearance and the motion branches, respectively. In the equations above, $w$ refers to the weights of the networks, $f1_{w}$ and $f2_{w}$ denote the networks of the appearance and motion branches, respectively. Moreover, SM stands for the function of producing Segmentation Mask (SM) by the Mask-RCNN, and OFM refers to the function that produces Optical Flow Magnitude Map (OFM).

\begin{figure}[t]
  \centering
   \includegraphics[width=0.4\linewidth]{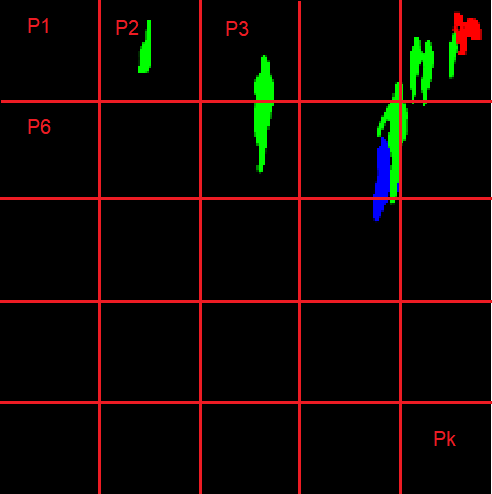}

   \caption{Image patches for the patch-level loss.}
   
   \label{fig:patch}
\end{figure}

\subsection{Inference}
Fig. 1 illustrates the inference phase. In this stage, separately for each branch, we compare the outputs of the trained networks with their expectations (target images) and compute their differences (i.e., the sum of the pixel-wise differences for all corresponding pixels) as the anomaly score. To this end, we calculate the MSE loss between the output and the target image and use it as the anomaly score. For an input frame, if the anomaly score of any branch passes its set threshold, the frame is labelled as an anomaly from an appearance or motion viewpoint. In all, we label a frame as an anomaly if any of the branch anomaly scores pass its threshold.

\subsection{Refinement}
As studied in \cite{mohammad1} one of the drawbacks of the existing methods is that their anomaly scores are affected by the number of foreground objects. In other words, more objects in a frame produce a higher anomaly score. Although normal objects produce a lower reconstruction/prediction error, a high number of normal foreground objects may result in a higher total anomaly score and have the same effect as an abnormal object. Analyzing the anomaly maps produced by our method and also the other methods, we found out that the anomaly activations of normal objects are non-condensing, and few steps of erosion and dilation morphological operations alleviate the problem without affecting the activations of the anomalies considerably. For this purpose, we applied few steps of the erosion and dilation operations respectively on the anomaly map to tackle the mentioned problem of non-invariance to the number of objects.

\subsection{Temporal denoising}
Methods based on object detection or segmentation may fail in detecting some objects in some frames which can lead to a sudden change in the anomaly score. On the other hand, considering the frame rates of the videos, adjacent frames are quite similar and are expected to produce similar anomaly scores. Hence, based on this assumption, we apply the Savitzky–Golay filter on the anomaly scores of the frames  \cite{kasun1}, in order to smooth the anomaly scores and remove the noise.

\section{Experiments and results}
\label{sec:experiments}

In this section, we evaluate the performance of our proposed method on three benchmarks (the ShanghaiTech \cite{liu1}, the UCSD-Ped1, and the UCSD-Ped2 \cite{mahadevan1}). The notions of normality and anomaly are the same in these datasets, and they are all suitable for semi-supervised training, as they contain normal frames in their training sub-sets. In addition to quantitative results, we provide numerous qualitative results to illustrate the activations in the anomaly maps which supports the explainability of our method. The details of the datasets and the experiments are provided in the following sub-sections.
\subsection{Datasets}

As mentioned, we evaluate the effectiveness of the proposed method on three benchmarks (ShanghaiTech, Ped1 and Ped2). These datasets are independently used both for training and testing. The ShanghaiTech dataset is captured in a university campus where walking pedestrians is normal. It contains different types of anomalies, such as abnormal objects (e.g., car, motorcycle, etc.) and abnormal motions (e.g., running, chasing, etc.). This dataset comes along with both pixel-level and frame-level annotations. The most distinctive features of this dataset are: 1) multiple scenes and view angles. 2) complex lighting conditions. In the UCSD-Ped1 and Ped2 datasets, the normal scenes include people walking in the walkways, while anomalies are due to the presence of unexpected objects in the scene (such as carts, bicycles, skateboards, etc.) or different motion patterns (skateboard riding, etc.). Most noticeable challenges of these datasets are: 1) comparatively low resolution and grayscale, which leads to difficulty recognizing the objects. 2) size of the people may change considerably regarding their distance from the camera. 3) considerable camera shakes in some frames. It is worth mentioning that in the ground-truth annotations of the mentioned datasets, a frame is annotated as an anomaly if it contains an anomalous object or motion.

\subsection{Evaluation metric}

We use the frame-level Area Under Curve (AUC) score as the accuracy metric in our experiments. By applying different thresholds on the anomaly score and gaining different True Positive Rates (TPR) and False Positive Rates (FPR), the Receiver Operation Characteristic (ROC) is plotted, and the Area Under the Curve is calculated to evaluate the performance of the method and consequently to compare it with state-of-the-art methods. A higher AUC shows a better performance. 
\subsection{Implementation details}

In our experiments all frames are resized to $224*224$ for both the appearance and motion branches. Besides, two consecutive frames$(\Delta t=1)$ are utilized to generate the optical flow target maps simply using the Farneback algorithm in the OpenCV library. To train the networks, we use the Adam optimizer to optimize the parameters. The learning rate of the training starts at 0.005; however, it is halved every 10 epochs. We have also set the K (number of the patches in the patch-level training) to 9 in our implementation. We also got the best performance for the refinement step, by applying one step of each erosion and dilation operations, respectively, and with the filter size set to 3.


\subsection{Qualitative analysis}
In this sub-section, we provide anomaly maps to visually validate the effectiveness of the contributions, and to demonstrate that our method focuses on the right aspect (class or the motion of the object) and the right place (object's position) to make its decision. As can be noticed in the anomaly maps, the major and higher activations are generated by anomalies which supports the explainability of the detections by our VAD method.

Fig. 5 and Fig. 6 show some qualitative results of the method on the Shanghai-Tech and the Ped2 datasets. These figures illustrate the performances of both the appearance (in colour images) and motion branches (in gray-scale images). For each sample frame, the following data are presented: the original input frame, the target map (expected output), the estimated output (output of the network), the anomaly map (difference between the expected and estimated outputs), and finally, the post-processed anomaly map. As shown in the figures, the networks of each branch which are trained on the normal frames, have learned normal appearances and motions so that for the normals, the networks produce outputs quite similar to their expected outputs (Semantic segmentation
map and optical flow magnitude map of the frame,
respectively for the appearance and the motion branches). However, the networks do not show the same behavior for anomalous objects. For the abnormal classes of objects, the appearance network either recognizes them as one of the previously seen objects (mostly as pedestrians) or detects them (or some parts of them) as background. Hence, the difference between the estimated and expected appearance generates a significant anomaly score in that position, in both cases. The motion network also estimates a motion quite close to its target motion map for a normal object, leading to a low anomaly score. However, for abnormal motions (usually faster motions), the estimation is close to the normal motions (i.e., different from their actual motions) and thus produce a big difference (i.e., higher anomaly score).

The calculated anomaly maps mainly consist of larger and dense activations at the position of the abnormal objects, and also some non-condensing activations (with a weak certainty) in some points of the normal objects. Even though the anomaly score is mainly affected by the anomalies in the raw anomaly map, the morphological post-processing operations decrease the effect of normals by eliminating non-condensing pixels. Our experimental results show that morphological refinement is more effective in the Shanghai dataset. By qualitatively analyzing the results, it is assumed that the difference in the performance is mainly due to the low resolution and the small size of the objects in UCSD datasets.

\begin{figure}[t]
  \centering
   \includegraphics[width=0.8\linewidth]{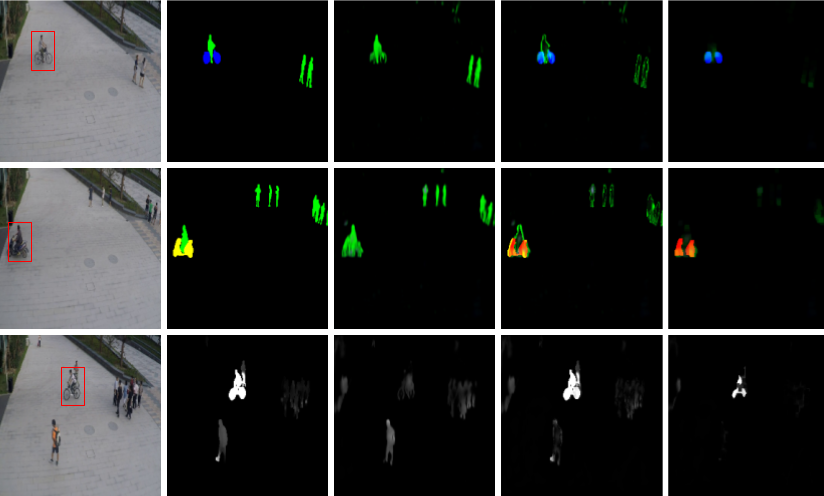}

   \caption{Qualitative results for the ShanghaiTech dataset. First two rows show the results for the appearance branch and the last row for the motion branch. Starting from left, columns show: input frame, target output, estimated output, anomaly map (difference of target and estimated images) and post-processed anomaly map. Red boxes indicate the ground-truth anomalies. Best viewed in color.}
   
   \label{fig:erod_sh}
\end{figure}

\begin{figure}[t]
  \centering
   \includegraphics[width=0.85\linewidth]{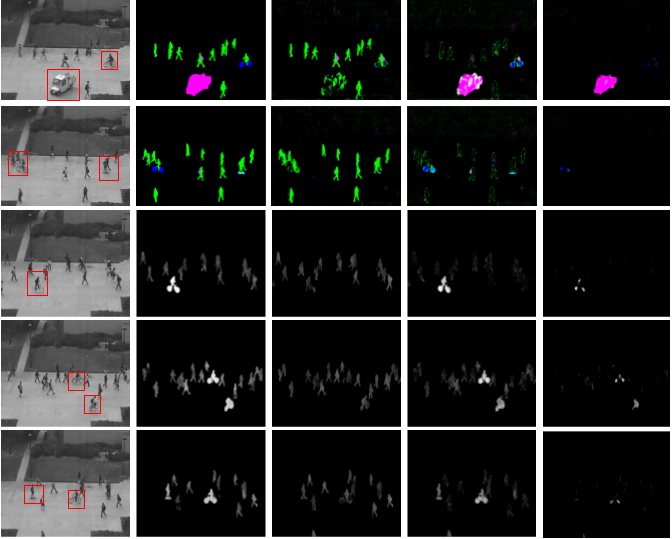}

   \caption{Qualitative results for the UCSD-Ped2 dataset. First two rows show the results for the appearance branch and the last three rows for the motion branch. Starting from left, columns show: input frame, target output, estimated output, anomaly map (difference of target and estimated images) and post-processed anomaly map. Best viewed in color.}
   
   \label{fig:erod_us}
\end{figure}

We also qualitatively evaluated the performance of the method in making association between normal objects and their motions. Fig. 7 shows how our model effectively learns to make a correspondence between an object’s location and its motion. Besides, it learns the association between object parts and their motion. For example, the output results show that the network has learned to estimate a larger motion for the legs compared to other body parts. Moreover, the network predicts a larger motion for the objects near the camera.

\begin{figure}[t]
  \centering
   \includegraphics[width=0.5\linewidth]{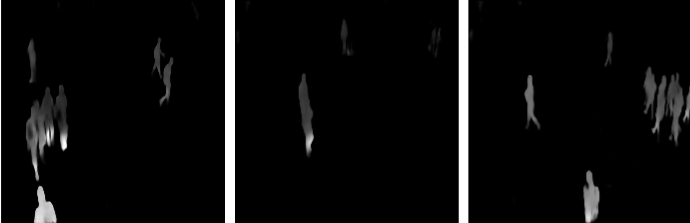}

   \caption{Normal motions estimated by the motion network for different locations or different object parts.}
   
   \label{fig:dist}
\end{figure}

Our experiments also show that most of the false positive detections are due to the detection of a normal object by one network (either the image translator or Mask-RCNN) and the misdetection by the other one. This produces a higher difference between the estimated and the expected outputs for that object (i.e., a large activation in the anomaly map). Our experiments show that the mentioned misdetection is mainly because of few factors such as: low resolution of the input frame, very small size of the objects, reflection of the objects in mirror, and camouflage with the background. Moreover, by analyzing the results we found that the appearance network (the image translator) outperforms the Mask-RCNN in segmenting the normal objects, and mostly the pre-mentioned failure (i.e., failure of one network in detecting normal objects) occurs in the Mask-RCNN.

Our method assumes that the appearance-based anomalies are due to the unseen object classes. Hence we do not expect good performance on the cases in which anomalies are due to the low-level image features (e.g., defect detection on the surface). Moreover, in order to increase the performance of the motion branch in making the correspondence between objects and their motion, we concentrate on the magnitude of motion which may decrease the performance for cases in which anomalies are due to the direction.


\subsection{Comparison with the state-of-the-art methods and ablation study}

In Table {I}, the effectiveness of our proposed method is compared with state-of-the-art works. The results show that our proposed method outperforms the other methods (considering AUC), for the ShanghaiTech, and it shows competitive results with others, for the UCSD datasets.

\begin{table}
  \centering
  \caption{Comparison between VAD methods. Best-performing method is denoted in boldface.}
  \label{tab:Aucs}
  \scalebox{0.95}{
  \begin{tabular}{@{}llll@{}}
    \toprule
    Method & Ped1 & Ped2 & ShanghaiTech\\
    \midrule
    
    Mahdyar et al. \cite{ravanbakhsh1} & \textbf{97.4} & 93.5 & N/A \\
    Hasan et al. \cite{hasan1} & 75.0 & 85.0 & 60.9 \\
    Chong et al. \cite{chong1} & 89.9 & 87.4 & N/A \\
    Liu et al. \cite{liu1} & 83.1 & 95.4 & 72.8 \\
    Park et al. \cite{park1} & N/A & 97.0 & 70.5 \\
    Nguyen et al. \cite{trong1} & N/A & 96.2 & N/A \\
    Hui et al. \cite{LV1} & 85.1 & 96.9 & 73.8 \\
    Yu et al. \cite{yu1} & N/A & 95.0 & 73.0 \\
    Ionescu et al. \cite{ionesco1} & N/A & \textbf{97.8} & 84.9 \\
    Our’s & 88.61 & 97.76 & \textbf{86.18} \\
    
    \bottomrule
  \end{tabular}}
  
\end{table}



\begin{table}
  \centering
  \caption{Ablation Study: comparing 6 cases.}
  \label{tab:ablation}
  \scalebox{0.75}{
  \begin{tabular}{@{}lllllll@{}}
    \toprule
    Item & C1 & C2 & C3 & C4 & C5 & C6\\
    \midrule
    Pre-trained encoder & \checkmark & - & \checkmark & \checkmark & \checkmark & \checkmark\\
    Masking & \checkmark & \checkmark & - & \checkmark & \checkmark & \checkmark\\
    Refinement step & \checkmark & \checkmark & \checkmark & - & \checkmark & \checkmark\\
    Appearance branch & \checkmark & \checkmark & \checkmark & \checkmark & - & \checkmark\\
    Motion branch & \checkmark & \checkmark & \checkmark & \checkmark & \checkmark & -\\
    \midrule
    AUC & \textbf{85.52} & 84.32 & 80.12 & 85.10 & 69.99 & 76.33\\
    \bottomrule
  \end{tabular}}
\end{table}

Table {II} provides results for an ablation study we performed on ShanghaiTech to analyze the contribution of individual components in the proposed method. The results show that using a pre-trained U-net and the masking stage have the most positive impact on the performance. The higher effect of the masking stage is likely because of camera shakes in numerous frames which produces background noise in motion maps and consequently generating a high false activations in the anomaly map. Moreover, the results demonstrate that the appearance and motion branches act as complementary to each other and their combination considerably improves the performance (15.53\% compared to only using the motion branch and 9.19\% compared to only using the appearance branch). 

Table {III} shows the improvements of the performance (AUC) by the temporal denoising step. Based on numerous experiment on the datasets we got the best results with window-length and the poly-order (hyper-parameters of the filter) equal to 41, 1, respectively. We assume that these numbers are achieved in connection with the frame rate of the video clips. As expected, temporal denoising step is more effective on UCSD datasets, since the Mask-RCNN shows comparatively more detection failures in them because of their lower resolutions.

\begin{table}
  \centering
  \caption{Performance (AUC) improvement by temporal denoising.}
  \scalebox{0.80}{
  \begin{tabular}{@{}llllll@{}}
    \toprule
    Dataset & ShanghaiTech & UCSD-Ped2 & UCSD-Ped1 \\
    \midrule
    W/O denoising  & 85.52  & 96.19 & 86.81 \\
    W denoising & 86.18 & 97.76  & 88.61 \\

    \bottomrule
  \end{tabular}}
  
\end{table}
\subsection{Visualizing frame anomaly score}

Fig. 8 shows the anomaly scores produced by the appearance network for two test clips. The figure shows that the anomaly score experiences a big rise for the frames containing an abnormal object class. Fig. 9 on the other hand, shows the performance of both appearance and motion streams in detecting anomalies by their corresponding anomaly scores. During the frames in which a bicycle passes, the anomaly scores of both streams observe a significant rise. Finally, Fig. 10 shows the generalization capability of the trained method (the appearance branch here as a sample) to the other scenes. This figure consists of the produced anomaly scores for a test clip (Ped2-test004) that is produced by networks trained on the Ped2, Ped1, and Shanghai-Tech, respectively. 

\begin{figure}[t]
  \centering
   \includegraphics[width=0.85\linewidth]{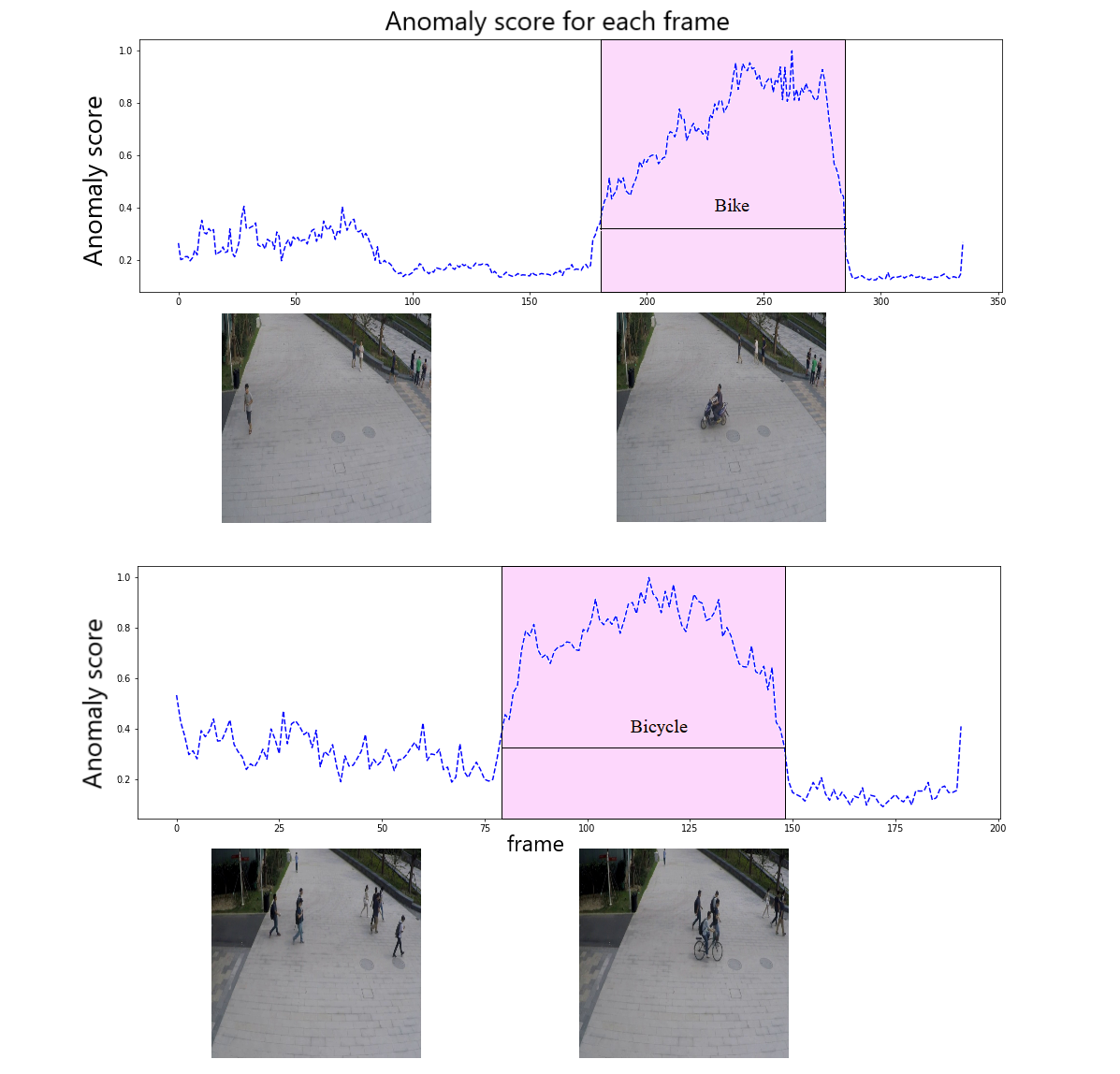}

   \caption{Anomaly score of frames, produced by the appearance branch, for two different clips. The pink highlight shows anomaly intervals.}
   
   \label{fig:score1}
\end{figure}

\begin{figure}[t]
  \centering
   \includegraphics[width=1\linewidth]{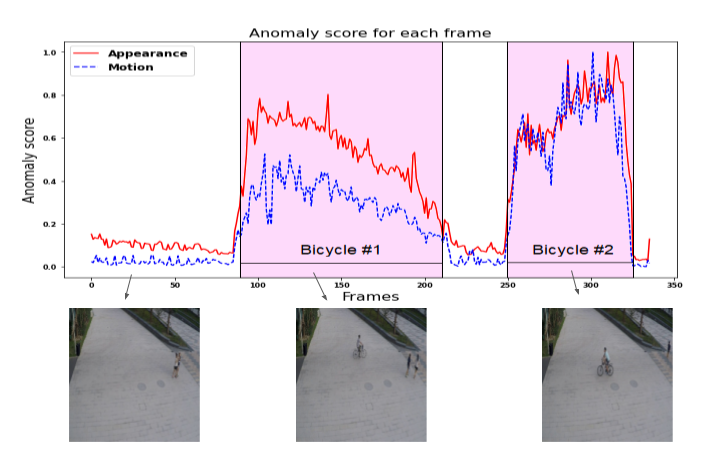}

   \caption{Anomaly score of frames from a sample clip. Red curve: anomaly score calculated by appearance branch. Blue curve: anomaly score calculated by motion branch. The pink highlight shows anomaly intervals.}
   
   \label{fig:score2}
\end{figure}

\begin{figure}[t]
  \centering
   \includegraphics[width=0.9\linewidth]{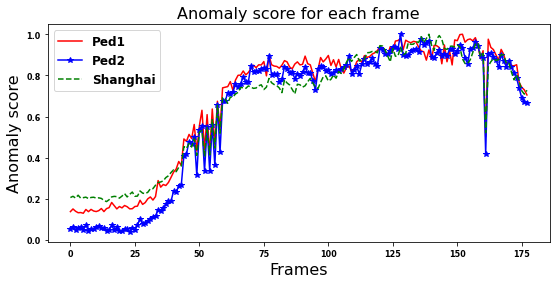}

   \caption{Generalization capability of the appearance branch. Blue-star: Anomaly score of a sample UCSD-Ped2 clip calculated by our network trained on the UCSD-Ped2 dataset. Red: Trained on the Ped1. Green-dash: Trained on the ShangahiTech dataset.}
   
   \label{fig:gen}
\end{figure}

\section{Conclusion}
\label{sec:conc}
We propose a novel object class aware VAD detection method which combines two complementary branches to detect anomalies in video. Both branches leverage the image translation task, to make an association between an input frame and its corresponding target images and consequently to learn normal patterns. The networks of the branches (U-Net) trained on normal frames, fail to make a correct correspondence between an abnormal input frame and its target images, in the inference stage which results in a larger activation in the anomaly map. Competitive quantitative results with state-of-the-art methods on benchmark datasets show good performance, and the qualitative results demonstrate the high explainability of our proposed method.





%
\bibliographystyle{IEEEtran}

\bibliography{IEEEabrv,mrv1}

\begin{thebibliography}{10}
\providecommand{\url}[1]{#1}
\csname url@samestyle\endcsname
\providecommand{\newblock}{\relax}
\providecommand{\bibinfo}[2]{#2}
\providecommand{\BIBentrySTDinterwordspacing}{\spaceskip=0pt\relax}
\providecommand{\BIBentryALTinterwordstretchfactor}{4}
\providecommand{\BIBentryALTinterwordspacing}{\spaceskip=\fontdimen2\font plus
\BIBentryALTinterwordstretchfactor\fontdimen3\font minus
  \fontdimen4\font\relax}
\providecommand{\BIBforeignlanguage}[2]{{%
\expandafter\ifx\csname l@#1\endcsname\relax
\typeout{** WARNING: IEEEtran.bst: No hyphenation pattern has been}%
\typeout{** loaded for the language `#1'. Using the pattern for}%
\typeout{** the default language instead.}%
\else
\language=\csname l@#1\endcsname
\fi
#2}}
\providecommand{\BIBdecl}{\relax}
\BIBdecl

\bibitem{chandola1}
V.~Chandola, A.~Banerjee, and V.~Kumar, ``Anomaly detection: A survey,''
  \emph{ACM Comput. Surv.}, vol.~41, no.~3, Jul. 2009.

\bibitem{liu1}
W.~Liu, D.~L. W.~Luo, and S.~Gao, ``Future frame prediction for anomaly
  detection -- a new baseline,'' in \emph{2018 IEEE Conference on Computer
  Vision and Pattern Recognition (CVPR)}, 2018.

\bibitem{hasan1}
M.~Hasan, J.~Choi, J.~Neumann, A.~K. Roy-Chowdhury, and L.~S. Davis, ``Learning
  temporal regularity in video sequences,'' in \emph{2016 IEEE Conference on
  Computer Vision and Pattern Recognition (CVPR)}, 2016, pp. 733--742.

\bibitem{chong1}
Y.~S. Chong, Y.~H. Tay, F.~Cong, A.~Leung, and Q.~Wei, ``Abnormal event
  detection in videos using spatiotemporal autoencoder,'' in \emph{Advances in
  Neural Networks - ISNN 2017}.\hskip 1em plus 0.5em minus 0.4em\relax Cham:
  Springer International Publishing, 2017, pp. 189--196.

\bibitem{abati1}
D.~Abati, A.~Porrello, S.~Calderara, and R.~Cucchiara, ``Latent space
  autoregression for novelty detection,'' in \emph{2019 IEEE/CVF Conference on
  Computer Vision and Pattern Recognition (CVPR)}, 2019, pp. 481--490.

\bibitem{ionesco1}
R.~T. Ionescu, F.~Khan, M.~Georgescu, and L.~Shao, ``Object-centric
  auto-encoders and dummy anomalies for abnormal event detection in video,'' in
  \emph{2019 IEEE/CVF Conference on Computer Vision and Pattern Recognition
  (CVPR)}, 06 2019, pp. 7834--7843.

\bibitem{doshi1}
K.~Doshi and Y.~Yilmaz, ``Any-shot sequential anomaly detection in surveillance
  videos,'' in \emph{2020 IEEE/CVF Conference on Computer Vision and Pattern
  Recognition Workshops (CVPRW)}, 2020, pp. 4037--4042.

\bibitem{kasun1}
W.~Chongke, S.~Sicong, T.~Cihan, S.~Pratik, and H.~Salim, ``An explainable and
  efficient deep learning framework for video anomaly detection,'' in
  \emph{2021 Cluster computing}, 2021, pp. 1--23.

\bibitem{samet1}
S.~Akçay, A.~Atapour-Abarghouei, and T.~P. Breckon, ``Skip-ganomaly: Skip
  connected and adversarially trained encoder-decoder anomaly detection,'' in
  \emph{2019 International Joint Conference on Neural Networks (IJCNN)}, 2019,
  pp. 1--8.

\bibitem{ravanbakhsh1}
M.~Ravanbakhsh, M.~Nabi, E.~Sangineto, L.~Marcenaro, C.~S. Regazzoni, and
  N.~Sebe, ``Abnormal event detection in videos using generative adversarial
  nets,'' \emph{2017 IEEE International Conference on Image Processing (ICIP)},
  pp. 1577--1581, 2017.

\bibitem{park1}
H.~Park, J.~Noh, and B.~Ham, ``Learning memory-guided normality for anomaly
  detection,'' \emph{2020 IEEE/CVF Conference on Computer Vision and Pattern
  Recognition (CVPR)}, pp. 14\,360--14\,369, 2020.

\bibitem{chen1}
D.~Chen, P.~Wang, L.~Yue, Y.~Zhang, and T.~Jia, ``Anomaly detection in
  surveillance video based on bidirectional prediction,'' \emph{Image and
  Vision Computing}, vol.~98, p. 103915, 2020.

\bibitem{pankaj1}
P.~R. Roy, G.-A. Bilodeau, and L.~Seoud, ``Local anomaly detection in videos
  using object-centric adversarial learning,'' in \emph{Pattern Recognition.
  ICPR International Workshops and Challenges}.\hskip 1em plus 0.5em minus
  0.4em\relax Cham: Springer International Publishing, 2021, pp. 219--234.

\bibitem{yu1}
Y.~Zhang, X.~Nie, R.~He, M.~Chen, and Y.~Yin, ``Normality learning in
  multispace for video anomaly detection,'' \emph{IEEE Transactions on Circuits
  and Systems for Video Technology}, vol.~31, no.~9, pp. 3694--3706, 2021.

\bibitem{zhao1}
Y.~Zhao, B.~Deng, C.~Shen, Y.~Liu, H.~Lu, and X.~Hua, ``Spatio-temporal
  autoencoder for video anomaly detection,'' \emph{Proceedings of the 25th ACM
  international conference on Multimedia}, 2017.

\bibitem{morais1}
R.~Morais, V.~Le, T.~Tran, B.~Saha, M.~Mansour, and S.~Venkatesh, ``Learning
  regularity in skeleton trajectories for anomaly detection in videos,'' in
  \emph{2019 IEEE/CVF Conference on Computer Vision and Pattern Recognition
  (CVPR)}, 2019, pp. 11\,988--11\,996.

\bibitem{mohammad1}
M.~Baradaran and R.~Bergevin, ``A critical study on the recent deep learning
  based semi-supervised video anomaly detection methods,'' \emph{arXiv preprint
  arXiv:2111.01604}, 2021, unpublished.

\bibitem{krzystof1}
K.~Lis, K.~Nakka, P.~Fua, and M.~Salzmann, ``Detecting the unexpected via image
  resynthesis,'' in \emph{2019 IEEE/CVF International Conference on Computer
  Vision (ICCV)}, 10 2019, pp. 2152--2161.

\bibitem{biase1}
G.~D. Biase, H.~Blum, R.~Y. Siegwart, and C.~Cadena, ``Pixel-wise anomaly
  detection in complex driving scenes,'' in \emph{CVPR}, 2021.

\bibitem{trong1}
T.~N. Nguyen and J.~Meunier, ``Anomaly detection in video sequence with
  appearance-motion correspondence,'' in \emph{2019 IEEE/CVF International
  Conference on Computer Vision (ICCV)}, 2019, pp. 1273--1283.

\bibitem{yang1}
Y.~Yang, D.~Zhan, F.~Yang, X.-D. Zhou, Y.~Yan, and Y.~Wang, ``Improving video
  anomaly detection performance with patch-level loss and segmentation map,''
  in \emph{2020 IEEE 6th International Conference on Computer and
  Communications (ICCC)}, 2020, pp. 1832--1839.

\bibitem{mahadevan1}
V.~Mahadevan, W.~Li, V.~Bhalodia, and N.~Vasconcelos, ``Anomaly detection in
  crowded scenes,'' in \emph{2010 IEEE Computer Society Conference on Computer
  Vision and Pattern Recognition}, 2010, pp. 1975--1981.

\bibitem{LV1}
H.~Lv, C.~Chen, Z.~Cui, C.~Xu, Y.~Li, and J.~Yang, ``Learning normal dynamics
  in videos with meta prototype network,'' in \emph{Proceedings of the IEEE/CVF
  Conference on Computer Vision and Pattern Recognition (CVPR)}, June 2021, pp.
  15\,425--15\,434.

\end{thebibliography}

\end{document}